\documentclass[sigconf]{acmart}

\usepackage{booktabs} 
\usepackage{amsmath}
\usepackage{algorithm}
\usepackage{algpseudocode}
\usepackage{bbm}
\usepackage{bm}

\makeatletter
\def\BState{\State\hskip-\ALG@thistlm}
\makeatother 

\usepackage{hyperref}
\usepackage{breakurl}
\usepackage{multirow}

\setcopyright{rightsretained} 
\acmConference[KDD'18 Deep Learning Day]{ACM KDD'18 Deep Learning Day}{August 2018}{London, UK}
\acmYear{2018}
\copyrightyear{2018}

\begin{document}

%

\title[Abstractive and Extractive Text Summarization]{Abstractive and Extractive Text Summarization using Document Context Vector and Recurrent Neural Networks}

\author{Chandra Khatri}
\authornote{Work was done while the author was at eBay Inc.}
\affiliation{%
  \institution{Amazon Lab126}
 Sunnyvale, California
}
\email{ckhatri@amazon.com}

\author{Gyanit Singh}
\affiliation{%
  \institution{eBay Inc.}
San Jose, California
}
\email{gysingh@ebay.com}

\author{Nish Parikh }
\authornote{Work was done while the author was at eBay Inc.}
\affiliation{%
  \institution{Google}
  Mountain View, California
  }
\email{nishparikh@google.com}


\begin{abstract}

Sequence to sequence (Seq2Seq) learning has recently been used for abstractive and extractive summarization. In current study, Seq2Seq models have been used for eBay product description summarization. We propose a novel Document-Context based Seq2Seq models using RNNs for abstractive and extractive summarizations. Intuitively, this is similar to humans reading the title, abstract or any other contextual information before reading the document. This gives humans a high-level idea of what the document is about. We use this idea and propose that Seq2Seq models should be started with contextual information at the first time-step of the input to obtain better summaries. In this manner, the output summaries are more document centric, than being generic, overcoming one of the major hurdles of using generative models. We generate document-context from user-behavior and seller provided information. We train and evaluate our models on human-extracted-golden-summaries. The document-contextual Seq2Seq models outperform standard Seq2Seq models. Moreover, generating human extracted summaries is prohibitively expensive to scale, we therefore propose a semi-supervised technique for extracting approximate summaries and using it for training Seq2Seq models at scale. Semi-supervised models are evaluated against human extracted summaries and are found to be of similar efficacy. We provide side by side comparison for abstractive and extractive summarizers (contextual and non-contextual) on same evaluation dataset. Overall, we provide methodologies to use and evaluate the proposed techniques for large document summarization. Furthermore, we found these techniques to be highly effective, which is not the case with existing techniques.

\end{abstract}

%
%


%
%

%
%




\maketitle


\section{Introduction}
\begin{figure}
\centering
\includegraphics[height=1.4in]{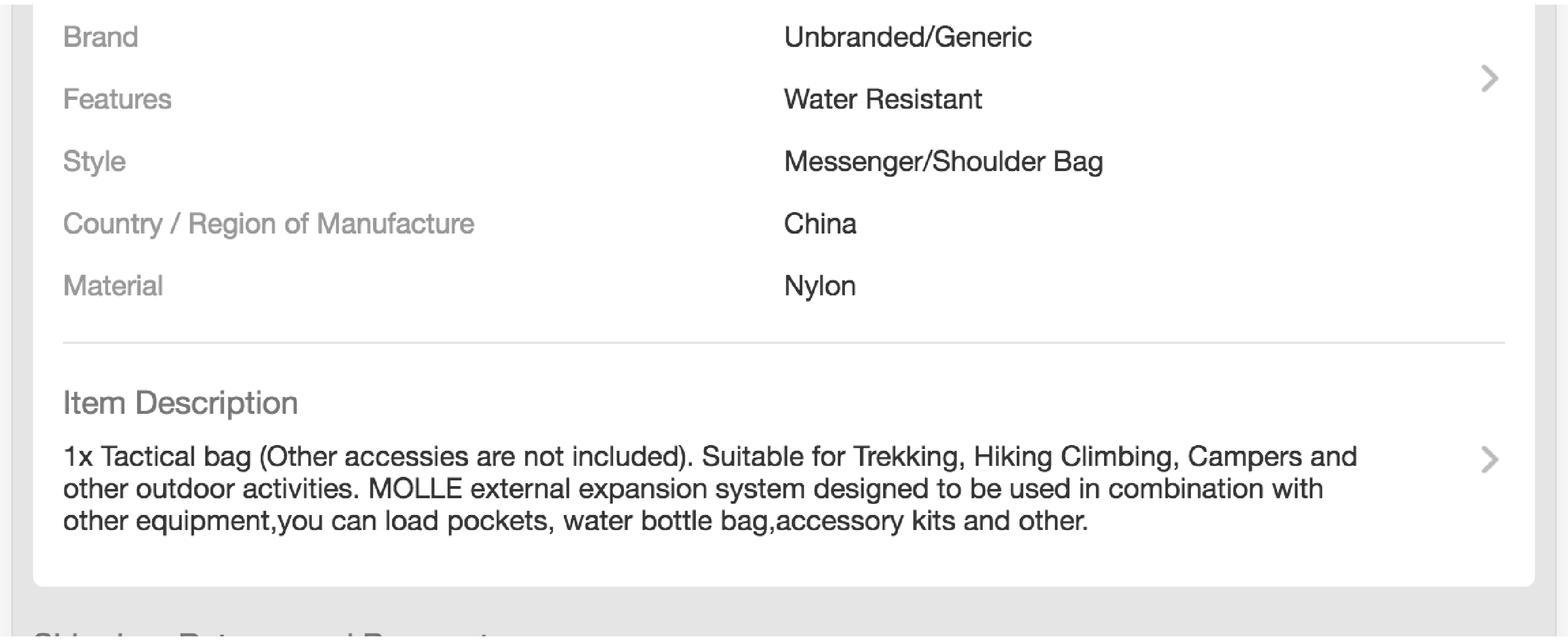}
\caption{Snapshot of a product snippet as it appears on the eBay mobile app for product titled \emph{"Mens Travel Hiking Military Tactical Army Camo Sling Backpack Chest Shoulder Bag". The full html description is hidden behind the click (arrow on "Item description") making the current page easy to consume and user friendly.} \label{fig:eg1}}
\end{figure}

Document summarization has its applications in almost all the domains of the Internet.
Search engines provide query and context specific summary snippets as a part of search experience, 
news websites use summaries to brief the articles, social media use them for content targeting 
while e-commerce websites use summaries for better browse experience through item or product highlights.
In this paper, we leverage data-sets from a popular B2C and C2C eBay. 
Presenting users with product summary decreases user's cognitive load to evaluate product's relevance to their purchase intent, leading to higher engagement and better browsing experience.
Furthermore, as the traffic on mobile sites and applications is increasing, item summaries become much more relevant. 
Due to limited real-estate available for mobile sites and design point of view it 
is more relevant to show item summaries than the entire HTML elements. 
Figure \ref{fig:eg1} depicts the picture of the snippet on eBay mobile application.


Text summarization techniques are either extractive or abstractive. 
In extraction, key sentences and objects are extracted without modifying the objects themselves. 
This is obtained by key-phrase or ad-hoc sentence extraction keeping the sentences intact \cite{badry2013text,barzilay1999using}. 
While abstraction involves paraphrasing the context-aware sentences after understanding the language \cite{baumel2017abstractive,rush2015neural}. 
Abstractive techniques generally requires large-scale data (documents and corresponding summaries) for training the models, for example news titles can be considered as summaries and the articles can be considered as the document. 
In many instances, while summarizing content generated by third party, web systems have a legal constraints on modification of the content.
In these cases, summaries are extracted than being generated. Since, majority of the content written for products in marketplace is provided by the sellers, 
the marketplace also has legal constraint about not revising the content.  

In this work, we propose a technique for generation and incorporating \emph{document context} in Seq2Seq based generative models for abstractive and extractive summarization.
Using document-context is akin to humans reading title and abstract to know the key-details before delving into the full document.
In this paper we describe techniques for generating context around documents using user behavior and other information provided by the document creators.
We show that RNNs for both abstraction and extraction, both benefit from feeding document context at the first time step of sequence to sequence learning. 
We evaluate the summaries generated by this methodology and found them to be more contextual to the document and preferred by humans. 

Human extracted summaries are used for training extractive models - like extractive-RNNs.
However, this approach for training is not scalable since extracting summaries for millions/hundreds of thousands of documents is prohibitively expensive.
Sequence to sequence model generally perform best with large scale data.
Therefore we proposed a novel approach using document-context for extracting approximate summaries in a semi-supervised fashion, which is used for large scale training.

Abstractive sequence to sequence models are generally trained on titles and subtitles. We adopt a similar approach with document-context, which helps us scaling the training. 
Furthermore, we perform a heuristics by using the trained model for re-ranking all the sentences (within the document) based on likelihood of a sentence in a document being a summary sentence during the inference. We adopted the idea from a state-of-the-art techniques for Automatic Speech Recognition, wherein an RNN is used for re-ranking the potential outcomes from an n-gram based language models \cite{xiong2017asr}.

Following are the main contributions of our work:-
\begin{itemize}
\item Obtaining the context vectors from documents, which can be used for extractive and abstractive summarization tasks.
\item Automatically extracting approximate summaries and creating training data to enable large scale semi-supervised learning for extractive summarization. This is shown to be competitive to supervised learning techniques.
\item Using RNN and CNN-RNN for extractive summarization.
\item Abstractive Seq2Seq summarization for large documents.
\item Using document-context vectors for improving the Seq2Seq  learning for text summarization. This novel approach is shown to beat the state of the art in a similar settings.
\item Comparing abstractive to extractive summarization under same setting of sequence to sequence learning.
\end{itemize}

\section{Previous Work} \label{sec:prevwork}
Several summarization techniques have been explored over the past decades \cite{steinberger2009evaluation,moawad2012semantic} following are some of the popular techniques:

\textbf{Surface level approaches} consider title words and cue-words (e.g. "important", "best" etc.) for extracting relevant sentences \cite{nastase2008topic,luhn1958automatic}. 

\textbf{Corpus based approaches} leverage structural distribution of words using internal or external corpus (e.g. WordNet \cite{Miller:1995:WLD:219717.219748}) for summarization. 

\textbf{Cohesion based approaches} considers cohesive relations between the concepts within the text such as (antonyms, repetitions, synonyms etc.) using Lexical Chains \cite{barzilay1999using} 

\textbf{Graph based approaches} are some of the most popular text summarization techniques. Each sentence in the text is represented as a vertex and a graph is constructed around all the sentences, where the edges correspond to the inter-connections between the sentences. 
LexRank \cite{erkan2004lexrank} and TextRank \cite{mihalcea2004textrank} are two such techniques. 

\textbf{Machine learning based approaches}: Document summarization can be converted to a supervised or semi-supervised learning problem. In supervised learning approaches, hints or clues such as key-phrases, topic words, blacklist words, are used to label the sentences as positive or negative classes or the sentences are manually tagged (which is not scalable). Once the labels are established, a binary classier can be trained for obtaining the scores or summary-likelihood score pertaining to each sentence. Several summarization techniques have been explored in the literature \cite{das2007survey} in this regard. However, classification based approaches generalize well, however they are not efficient in extracting document specific summaries. Training data for machine learning approach contains label of the sentences irrespective of the document. If the document level information is not provided then these approaches provide same prediction irrespective of the document. Providing document context in the models alleviates this problem, which is what is one of the contributions of this paper.

\textbf{Abstractive summarization} techniques are less prevalent in the literature than the extractive ones. It is much harder because it involves re-writing the sentences which if performed manually, is not scalable and requires natural language generation techniques. The two common abstraction techniques are structured and semantic \cite{moawad2012semantic,rush2015neural}, both of which mostly are either graph/tree based or ontology and rule (e.g. template) based. 

All the approaches mentioned above work well; however, they either face challenge towards scalability or large scale evaluation or do not generalize well. Due to complexity constraints, research to date has focused primarily on extractive methods, but due to advancements in Seq2Seq and Natural Language Generation techniques is making it possible to generate reasonable summaries for very short descriptions using Abstraction \cite{chopra2016abstractive,nallapati2016abstractive,nallapati2017abstractive,xu2017abstractive,baumel2017abstractive,li2017abstractive,khandelwalneural,lopyrev2015generating,rush2015neural,textsummarization}. \cite{paulus2017abstractive} used reinforcement learning while  \cite{gehring2017abstractive} used CNNs for abstractive summarization. Most of the current advancements are around short summaries from short documents. Furthermore, there does not exist any work where extractive and abstractive techniques have been compared in the similar setting using Seq2Seq approaches.

Seq2Seq techniques based approaches have been used to efficiently map the input sequences (description / document) to map output sequence (summary), however they require large amounts data. With several examples, model tends to learn the mapping between input sequence and output sequence and generate more efficient summaries corresponding to the input document. Moreover, it is found that Seq2Seq  models currently work well for smaller document summarizations (one-two lines of the document mapping to headlines/phrase representation) \cite{chopra2016abstractive,textsummarization}. Even though Seq2Seq models are providing benchmark results in Machine Translation and Speech Recognition tasks \cite{sutskever2011generating,cho2014learning,wiseman2016sequence} they have not yet performed well for summarization tasks, dialog systems and evaluation of dialog systems \cite{ram2017dialog, guo2017dialog, anu2017dialog} and are facing many challenges (e.g. summarizing long documents). For long sequence \cite{sutskever2011generating} depicted that reversing the source sentence provides better results. 

Architecturally our abstractive models are similar to \cite{chopra2016abstractive}, with a change in encoder being novel Document-Contextual-LSTM instead of simple attention-based mechanism. Document-Contextual-Vector is described in Section \ref{sec:doccontext}. Furthermore, unlike related previous work, models in current study are summarizing full/large documents (longer than 2500 characters) to generate relatively large summaries (800 characters) using Beam-search and vocabulary restrictions constraints. Furthermore, we have not found any literature where Seq2Seq based abstractive summarization is compared with related RNN/CNN variations for extractive summarization, which is novel addition of this paper. 
Khatri et. al. \cite{khatri2015algorithmic,khatri2016snippet,khatri2016snippet2} describes summarization for eCommerce setting.

\section{User actions to infer document context }\label{sec:doccontext}
Websites which host user generated documents for consumption by other users provide myriads of ways for document discovery.
For example, users on eCommerce website, discover relevant document to their intent via search, recommendation modules or various topical pages. 
Also, sellers or content creators while creating the document or product pages provide lot of metadata on the document, for example title, tags, categorical, etc.
In this section we describe how we use historical information from document creator and document consumer to generate the document context.
This context is then later combined with word embeddings obtained via Skip-Gram Model with Negative Sampling (SGNS) \cite{mikolov2013distributed} 
to generate what we call \emph{document context vectors} or DCV.
We also use document context to score sentences in the document for algorithmic labeling for large scale semi-supervised learning to obtain approximate summaries.
Stop words and high frequency words in vocabulary are not considered while obtaining the context words in the following subsections. 

Let $\mathbb{V}$ be the vocabulary of the corpora. We assume that stop words have been removed from the corpora to generate the vocabulary.
Also let $N$ be the vocabulary size i.e. $|\mathbb{V}| = N$.
For the document $d$, we generate three \emph{unit-vectors} namely $\widehat{C_s^d}$ ($s$ denotes seller), $\widehat{C_q^d}$ ($q$ denotes queries), 
$\widehat{C_b^d}$ ($b$ denotes browse). All of these vectors have dimension $N$, but for document $d$ most of the dimensions are 0. We later combine these vectors with SGNS vectors to make DCV. For brevity we drop the $d$ from superscript is rest of the section as vectors are being generated per document only.

\subsection{Document creators}
When a seller lists a product for sale on eBay they create a web-document for it, 
which includes title, key-value meta-data on fixed dimensions like condition, brand, size, color etc. 
They also type in verbose description containing images, videos and html. 
In addition to this they select a leaf in the eBay taxonomy, 
$C = \{l_1, l_2, \ldots, l_n\}$ where $l_i$ is taxon used in the taxonomy. 
The taxonomy used in eBay is a laminar family. 
That is given two taxa $l_1$ and $l_2$ either $l_1 \cap l_2 = \phi$ or $l_1 \subset l_2$ or $l_2 \subset l_1$. 
Seller has to attach a leaf in this taxonomy tree to the document. 

\begin{figure}
\centering
\includegraphics[height=0.8in]{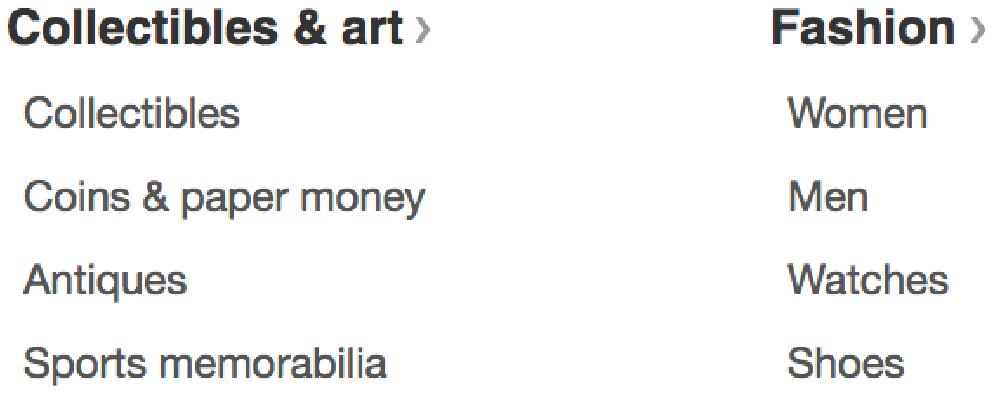}
\caption{ A snapshot of taxonomy used in a eBay. Here the top taxa Collectibles and Art are expanded
into lower taxa for example Antiques.\label{fig:taxa}}
\end{figure}

Figure \ref{fig:taxa} shows a small snapshot of eBay taxonomy. This taxonomy
tree is maintained and generated by domain experts and is of high quality. For example the document titled
\emph{"Mens Travel Hiking Military Tactical Army Camo Sling Backpack Chest Shoulder Bag"} is chosen to be put in the following leaf	
\emph{Sporting Goods > Outdoor Sports > Camping \& Hiking > Hiking Backpacks > Day Packs} by the seller.

Context vector $C_s$ is induced from seller provided metadata 
$m_s$=\emph{\{title, subtitle, taxon, other key-valued metadata such as brand, color etc.\}}. 
For $w \in \mathbb{V}$ let $C_s(w)$ be the value of the dimension $w$ in $C_s$. It is defined as follows,
\begin{align*}
    C_s(w) = 
\begin{cases}
    frequency(w) \text{ in } m_s,& \text{if } w \in \text{ seller metadata}\\
    0,              & \text{otherwise}
\end{cases}
\end{align*}

We then normlaize the vector $C_s$ to a unit vector $\widehat{C_s}$. We use the weight obtained for each context word later on to obtain  the document context vector for each document giving relevance to words based on weights. 
This technique is not limited to eBay or any eCommerce based pages only. Such an approach can be easily extended to the general textual documents available online in the form of webpages (e.g.: title, metadata, etc.) or articles (e.g.: title, sub-titles, abstract, keywords, category, etc.) 

\subsection{Document consumers}
Buyers discover relevant inventory via search, recommendations made to them or on topical pages 
( e.g. \burl{http://www.ebay.com/sch/Hiking-Backpacks-Bags/181378/bn_7407781/i.html}  $\&$ \burl{https://www.amazon.com/hiking-backpacks-and-bags/b?ie=UTF8&node=3400391} ) or from external sources like advertising etc. 
These discovery browse paths whether they are search trails or click trails \cite{singh2011user} that ends in a document are used to extract relevant words for the document.
One such discovery path is search. For example, when the user searches with the queries 
like \emph{"tactical sling backpack"}, \emph{"military backpack"} and lands on the document titled 
\emph{"Mens Travel Hiking Military Tactical Army Camo Sling Backpack Chest Shoulder Bag"} the 
words of the queries contains what the user thought was the most descriptive information from the document.
When such information is aggregated statistically cross large number of users it can provide great context for the document.

Let, $qset^d =\{$ queries used to discover document d$\}$ and $browse^d=\{$ titles of documents via which user discovered document $d$ (via recommendation or topical page) $\}$. For the dimension associated with word $w$ values in both vectors are defined as 
\begin{align*}
    C_q(w) = 
\begin{cases}
    frequency(w) \text{ in } qset^d,& \text{if } w \in qset^d\\
    0,              & \text{otherwise}
\end{cases} 
\\
C_b(w) = 
\begin{cases}
    frequency(w) \text{ in } browse^d,& \text{if } w \in browse^d\\
    0,              & \text{otherwise}
\end{cases}
\end{align*}
We then normalize the vector $C_b$, $C_q$ to a unit vector $\widehat{C_q}$, $\widehat{C_b}$. Similar to the weights obtained for the context words from document creators, we obtain the weights for words provided by the document consumers/readers. Basically, there are significant information provided by the readers based on their experience and behavior. We can leverage the information provided by the readers to further obtain more context words and corresponding weights, which is later on used to obtain the document context vector.

\subsection{Document Context Vectors} \label{subsection:vd}
For a document $d$ we defined three unit vectors above $\widehat{C_s^d}$ ($s$ denotes seller), $\widehat{C_q^d}$ ($q$ denotes queries), 
$\widehat{C_b^d}$ ($b$ denotes browse).
We combine these three to form a cumulative context unit vector $C^d$ by adding these three vector.
\[
C^d = \beta_s * \widehat{C_s^d} + \beta_q * \widehat{C_q^d} + \beta_b * \widehat{C_b^d}
\]
We used $\beta_s=1, \beta_q=1$ and $\beta_b=1$. These parameters may be fine tuned based on historical demand, 
which may be estimated by ratio of traffic volumes in different channels 
or by expert set pre-existing priors for importance of various channels of traffic.
We then re-weight value for each word dimension ($w$) in $C^d$ by its \emph{idf} to make $C^d_{idf}$. 
More precisely value for dimension for word $w$ is defined as 

\begin{align*}
    C^d_{idf}(w) = C^d(w) * idf(w) \text{\;\;for word } w
\end{align*}

We combine context with word embeddings obtained via SGNS to generate a vector per document that we call
$v_d$. Let $M_{SGNS}$ be the matrix of dimension $Nxk$ where $k$ is the dimension of word embeddings and $N$ is the vocabulary size of the corpora.
Row for word $w$ in $M_{SGNS}$ is the word embedding of the word $w$. We define $v_d$ as 

\begin{equation}
\label{equation:vd}
v_d = C^d_{idf} * M_{SGNS}
\end{equation}

Note that dimension of $v_d$ is $1xk$ just like word embeddings.

\subsection{Scoring Sentences Using Document Context\label{section:scoring}}
For a document $d$ let's say it contains $Sentneces_d = \{$ set of sentences in $d\}$. We describe how to use document context to score 
these sentences. These score are used for generating algorithmic labels for large-scale semi-supervised learning to obtain approximate summaries. 
For more details see Section \ref{section:datasets}. For a sentence $S \in Sentneces_d$, where $S = w_1 w_2 \ldots w_k$.

\begin{equation*}
\label{eqn:sentencescore}
S_{score} = \sum_{w \in S} v_d(w)
\end{equation*}
 
The above score corresponds to the weighted sum for the words in the sentence. The weight for the words incorporate frequencies in seller provided metadata, buyer's search history, browse history (leading to discovery of the said product) and also the Inverse Document Frequence (IDF) score of that word. The IDF score for each word is simply the inverse of documents containing this word across all the documents at eBay. 
The IDF score corresponds to the topical or document level relevance for each word in the vocabulary. Highly common words such as stop words ("the", "and", etc. have nearly zero IDF score). Using the score above, we obtain the score of each sentence. We use the sentence score to rank each sentence within the document and select top k (till we reach 800 characters) sentences as the summary for a document/product. We did an A/B test to see if the summaries obtained using this technique correlates with user expectations and we observed statistically significant lift in sales and user-engagement. This implies that the summaries obtained using this semi-supervised approach are useful. We use this technique to generate the training data for supervised extraction based summarization techniques (e.g. RNN).

\section{Models}

This work primarily focuses on adding context as the initial state to RNNs for abstractive and extractive text summarizations and comparing it with various state-of-the art techniques. 
For extraction we use labeled supervised and semi-supervised data. For abstraction we use titles and subtitles for training the models.

\subsection{Nomenclature and Base Model}

Recurrent Neural Network is a type of neural network which is an extension to Feed Forward NN, 
with at least one feed-back connection, so activations can flow round in a loop \cite{sutskever2011generating,elman1990finding}.
Essentially, information from prior observation along with the current observations are used to make predictions.

The notations are borrowed from Sutskever et al. \cite{cho2014learning}. RNN computes an output sequence $(y_1,y_2,\ldots,y_T)$ for a given input sequence $(x_1,x_2,\ldots,x_T)$ corresponding to the following equations:
\begin{equation}
  h_t = sigm (W^{h_x} x_t + W^{h_h} h_{t-1}  )
\end{equation}
\begin{equation}
  y_t=W^{y_h} h_t
\end{equation}
   
This framework works when there is an alignment between input and output sequences that is the size of the input is same as size of the output. When the size of input and output sequences are different then two RNNs with encoding-decoding mechanisms are used ( Cho et al. \cite{cho2014learning}). Theoretically this framework should work however it is found that RNNs with encoding-decoding mechanisms find difficulties in mapping long sequences or when there are long term dependencies. Gated Recurrent Units (GRUs) \cite{chung2014empirical}  and Long Short Term Memory (LSTMs) ) \cite{song2016training} solve this problem by introducing gates into the network to prevent vanishing gradient problem associated with RNNs with long term dependencies.  

In current study, LSTMs are used for extractive and abstractive summarizations. In standard LSTM sequence of fixed length is passed as an input to be encoded into a fixed dimension vector $(\textit{v})$, which is then decoded into the output sequence of words. To summarize, LSTM estimates the following:

\begin{equation}
p(y_1,y_2, \ldots,y_{T'} | x_1,x_2, \ldots ,x_T) = \prod_{t=1}^{T'} p(y_t | v,y_1,y_2,\ldots,y_{t-1})
\end{equation}

In abstractive summarization, document can be fed as an input during training and the summaries can be fed as output. However, extractive RNN can be trained using standard supervised classification setup
by performing soft-max on the encoded layer. \cite{cheng2016neural}.

\subsection{Contextual Recurrent Neural Network}
\subsubsection{ \textbf{ Abstractive Contextual RNN (AC-RNN)}}
It is an RNN architecture wherein a document context vector as described in Section \ref{sec:doccontext} is passed as an input at first time step along with the document sequence in the encoder. The idea is that if a pre-learned document-context-vector $(v_d)$ is passed as an input at the beginning of the encoding stage, then the model not only converges faster but also learns the summaries corresponding to the document and not just the generic sequences. The basic idea is that if a reader is aware of the \emph{title} of a document or \emph{abstract} for a publication, then it provides a better understanding and a high-level interpretation of the document which makes the model to be able to provide more specific summaries corresponding to the documents. 
Therefore, document-context-vector $(v_d)$ at the first time-step essentially changes the encoding vector. 

An LSTM decoder with similar architecture mentioned in encoding have been used, however the input at time t=0 to the decoder is the vector obtained from encoder \cite{sutskever2014sequence,cho2014learning}. Unlike other decoding mechanisms, where output at time $t$ can be any word from the vocabulary, the output from the document vocabulary is considered during the time of prediction, making the inference faster. 

Sutskever et al. \cite{sutskever2014sequence} and \cite{wiseman2016sequence} proposed beam-search to obtain the most likely sentence in the machine translation task. 
However, we perform a heuristics by using the trained model for re-ranking all the sentences (within the document) based on likelihood of a sentence being a summary sentence during the inference. Where likelihood of any sentence is defined as likelihood of decoding the sentence given the encoded input. So, the abstractive model is used for extraction during the inference. We adopted the idea from a state-of-the-art techniques for Automatic Speech Recognition, wherein an RNN is used for re-ranking the potential outcomes from an n-gram based language model \cite{xiong2017asr}. We do this to address several issues: (a) Avoid generic and short output issues with sequence to sequence models, (b) Obtaining grammatically correct sentences for eBay users to avoid poor customer experience and (c) Avoiding legal push-backs from the sellers. 
\subsubsection{\textbf{ Extractive Contextual RNN (EC-RNN)}} \label{sec:ecrnn}
It consists of Encoder only. 
The encoder used in EC-RNN is the replica of encoder used in AC-RNN with document-context-vector $(v_d)$ as an input at time t=0 and embedding representation of the words is passed as input to the model. However, the output of encoder is used for binary classification (sentence being a summary sentence or a not) using softmax. 
Note that, each sentence starts with document-context-vector, therefore the classification of sentence happens given the context of the document and not just the sentence alone. In this way same sentence may be classified as summary sentence for one document but not for others. Furthermore, given a document-context-vector ($v_d$), it is the extra information provided by the sentence which differentiates it from the other sentences in the document, which is a major drawback of other state-of-the-art classification approaches wherein some sentences are always classified as a summary sentence. 

\subsection{Non-Contextual RNN Architectures}

In current setup, RNNs trained without document-context-vector are termed as non-contextual RNNs. Recently, several architectures have been proposed in this regards \cite{chopra2016abstractive,nallapati2016abstractive,khandelwalneural,lopyrev2015generating,rush2015neural,textsummarization}, however three main models which are explored in current study are:

\subsubsection{\textbf{Abstractive RNN (A-RNN)}}
Abstractive RNN is the traditional sequence to sequence model using LSTM suggested by \cite{sutskever2014sequence,cho2014learning}. The model is exactly similar to AC-RNN without the context as input at time t=0. The input at time t=0 in A-RNN is a token \emph{<start>}. A fixed sized input and output sequences are used for training by either curtailing or padding. 

\subsubsection{\textbf{Extractive RNN (E-RNN)}}
 RNN has been used for classification tasks and it generates state-of-the art results. Extractive RNN is non-context version of EC-RNN proposed in Section \ref{sec:ecrnn}. As mentioned before, embeddings for words are pre-calculated using Skip Gram with Negative Sampling (SGNS) technique and are used as input corresponding to each word in encoding layer. Features are extracted using embedding for classification task. 

\subsubsection{\textbf{Convolutional RNN (CNN-RNN)}}
Convolution based LSTM has performed extremely well in text classification tasks \cite{zhou2015c}. Furthermore, convolution attention-based encoder \cite{chopra2016abstractive} has been used for short summarization tasks. 
CNN-LSTM is used to classify the sentences with the same technique as E-RNN, however the difference is that CNN is used to extract sequences of higher-level phrase representations. As suggested by Zhou et al. \cite{zhou2015c} CNN-LSTM is able to capture both local features of phrases, global and temporal sentence semantics. CNNs with multiple filters, max-pooling and dropout are used to extract high-level phrase representations and then passed to LSTM for classification using Softmax.

\section{Datasets}
Table \ref{tab:datasetdesc} describes the distribution of a eBay description. Vocabulary size of our dataset is 768K words. Median document length is 346 and 54 words.
\begin{table}
\centering
\caption{Description distribution of a eBay products.}
\label{tab:datasetdesc}
\begin{tabular}{|c|c|} \hline
Total vocabulary size & 768,298 \\ \hline
Median document length & 346 characters \\ \hline
Median number of words & 54 \\ \hline
Median sentence length & 51 characters \\ \hline
Median Nbr of words in sentence & 8 \\ \hline
\end{tabular}
\end{table}


\subsection{Datasets} \label{section:datasets}
There are two kinds of datasets which are used in current study.

\textbf{Human Extracted Snippets}: 20,000 items/documents and corresponding details (titles, url, description) were provided to humans for extracting the summaries. 
The task was to extract and rank the sentences from the descriptions given eBay item url. 
5k items out of 20k items were used for evaluations (Golden Set) and 15,000 items were used for training different models.

\textbf{Semi-supervised Large Scale Summarization Approximation}: 
Sentences from 100,000 item-descriptions were extracted and ranked based on relevance towards the document context. 
Section \ref{section:scoring} provides information about how to rank a sentence in the order of relevance given the document and it's contextual details. 
Several techniques have been proposed by Shen et al., Lin et al. and \cite{gong2001generic,badry2013text} based on query, thematic similarity, topic signature and Latent Semantic Analysis . \cite{nastase2008topic} expanded the topics to Wikipedia and obtained best results on DUC summarization task. 

Similar to approaches mentioned above, we obtain the approximate summaries for eBay item descriptions. After an evaluation from a eBay reviewers, it is found that summaries generated using document-context based approach are of high quality and can be used for training the models. Given the quality of these summaries, eBay launched this feature on mobile applications and website. On an A/B test, it is found that showing summaries using this approach have a high monetary value when compared to not showing the summary-snippet. 
Models trained using semi-supervised approach will be evaluated on Golden test set to identify the relevance of this technique.

\subsection{Data Generation for Classification task}
EC-RNN, C-RNN, CNN-RNN and other classification based Extractive summarization techniques need the labeled data for training. For classification tasks, sentences which have blacklist terms are labeled as \emph{non-summary} class while the sentences scoring high on document-context metric are considered as positive sentences. Blacklist terms are the words and phrases which do not contain item/document level information and are frequently used at eBay such as "returns", "shipping","rate me 5 stars", etc. We obtained 700 terms using human curation and with statistical analysis of eBay item descriptions. For each description in 100,000 items, sentences are tagged as positive or negative. Sentences which were neither scoring high on document-context metric nor having blacklist terms were left out from being tagged. Since, the document-context sentence ranking is a new approach and is yet to be evaluated except using A/B test in production, the data is tagged for high-precision. This work is a step towards evaluating document-context based sentence ranking as well. 

\section{Architecture Details and Experimentation}
\subsection{Training Details and Model Architectures}
\begin{table} 
\centering
\caption{Parameter setting for Abstractive approaches like AC-RNN and A-RNN.}
\label{tab:params:abstractive}
\vspace{-0.2cm}
\begin{tabular}{|c|c|} \hline
Parameters & Value \\ \hline
Input description length & 50 words \\ \hline
Output summary length &  15 words \\ \hline
\multirow{2}{*}{Optimization Method} &  Stochastic Gradient Descent \\
 & with momentum \\ \hline
\multirow{2}{*}{Learning rate} & 0.1; reduced to half after \\
&  every third epoch. \\ \hline
Batch size &  128 \\ \hline
LSTM Parameters &  Uniform distribution from [-0.1, 0.1] \\  \hline
\end{tabular}
\end{table}

\begin{table} 
\centering
\caption{Parameter setting for Extractive approaches like EC-RNN and E-RNN.}
\label{tab:params:extractive}
\begin{tabular}{|c|c|} \hline
Parameters & Value \\ \hline
Maximum sentence length & 15 words\\ \hline
Optimization Method & Adam \\ \hline
Learning rate & 0.01 \\ \hline
Batch size & 256 \\ \hline
LSTM Parameters  & Uniform distribution from [-0.1, 0.1] \\ \hline
\end{tabular}
\end{table}

\begin{table}
\centering
\caption{Parameter setting for Convolutional RNN .}
\label{tab:params:CNNRNN}
\begin{tabular}{|c|c|} \hline
Parameters & Value \\ \hline
Maximum sentence length & 15 words \\ \hline
Dropout keep probability & 0.5 \\ \hline
Learning rate & 0.01 \\ \hline
Filter sizes for Convolution & 4 \\ \hline
Batch size & 128 \\ \hline
\multirow{2}{*}{CNN and LSTM Parameters}&Random Normal centered at 0\\
    &with standard deviation 0.1\\
    \hline
Max pool size & 4 \\ \hline
\end{tabular}
\end{table}

\begin{table*} 
\centering
\caption{Different type of experiments done and metrics used for those experiments.}
\label{tab:metric}
\begin{tabular}{|c|c|c|} \hline
Evaluation Setting & Description & Evaluation Metric \\ \hline
\multirow{2}{*}{Classification}  & Given a sentence, classify whether  &  \multirow{2}{*}{Accuracy, Precision, Recall, F-score}\\ 
& it is summary sentence or not. & \\\hline
\multirow{2}{*}{Similarity} & Given golden summaries, & ROUGE (ROUGE-1, ROUGE-2, ROUGE-L) \cite{lin2004rouge},  \\ 
& find the similarity score  & BLEU, TF-IDF Cosine Similarity, Topic Similarity\\ \hline
\multirow{2}{*}{Ranking} &  Given sentences, rank them by in & \multirow{2}{*}{NDCG,  Mean Average Precision} \\
& the order of relevance towards summary  & \\ \hline
\end{tabular}
\end{table*}

\textbf{Abstractive Context RNN (AC-RNN) and Abstractive RNN (A-RNN)} were trained using deep LSTM with 4 layers (as described in Sutskever et al. \cite{sutskever2014sequence}) with 1000 cells and 300 dimension word embeddings. Since, we wanted to find the relative difference after adding the context in RNNs for summarization task, we kept the same parameters for both the models. Parameters settings and model details which worked the best in our case are mentioned in Table \ref{tab:params:abstractive}.

\textbf{Extractive Context RNN (EC-RNN) and Extractive RNN (E-RNN)} were trained using two LSTM layers with 300 cells and 300-dimension word embedding. Since, we wanted to find the relative difference after adding the context in RNNs for summarization task, we kept the same parameters for both the models. Parameters settings and model details which worked the best in our case are mentioned in Table \ref{tab:params:extractive}.

\textbf{Convolutional RNN (CNN-RNN)} consists of two neural networks. CNN for high level phrase representations and then LSTMs for obtaining the temporal and sequential nature of the text. We used single layer Convolution with filter size equal to 4 and a Single layer LSTM with 300 cells and 300-dimension word embeddings were used. Parameters settings and model details which worked the best in our case are mentioned in Table \ref{tab:params:CNNRNN}.

\subsection{Experiments and Evaluation Metrics}
We split our 20K human extracted summaries dataset into 15K and 5K parts. We use 5K for all evaluations. We also used the data obtained using semi-supervised technique for training and evaluating the models. 
\textbf{Supervised} - we train our models on 15K human extracted summaries. These models are trained with three different settings of target summary lengths. We use 1 sentence, 3 sentence and 5 sentence summary lengths. \\
\textbf{Semi-supervised} - we use 100K documents and approximate summaries generated via document context for training models at scale. We use 5K human extracted summaries for evaluation purposes.
We also baseline with a fuzzy summarization strategy where random sentences are picked in the summary output.

\bgroup
\def\arraystretch{1}
\begin{table}
\centering
\caption{Results on supervised task using human extracted summaries - 15K
for training and 5K for evaluation. 2 variations
with target summary :- (a) 1 sentence long and (b) 3 sentences
long. Abstractive Context RNN (AC-RNN) performs best in all metrics for short summaries. Extractive Context RNN
(EC-RNN) performs best for longer summaries. Adding context as a whole created improvements
in RNN models. }
\label{tab:goldresult}
\begin{tabular}{|c|c|c|c|c|c|c|} \hline
\multirow{2}{*}{Model} & Token  & \multirow{2}{*}{$Rouge_1$} & \multirow{2}{*}{$Rouge_2$} & Rouge & \multirow{2}{*}{BLEU}  & Topic  \\ 
 & Sim. & & & -LCS &  & Sim. \\ \hline \hline
\multicolumn{7}{|c|}{Target snippet length = 3 sentence} \\ \hline
Fuzzy & 0.18 & 0.26 & 0.20 & 0.20 & 0.15 & 0.01 \\ \hline
E-RNN & 0.20 & 0.27 & \textbf{0.37} & 0.27 & 0.16 & 0.02 \\ \hline
EC-RNN & \textbf{0.26} & \textbf{0.39} & 0.22 & \textbf{0.30} & \textbf{0.25} & 0.02 \\ \hline
CNN-RNN & 0.21 & 0.29 & 0.21 & 0.27 & 0.21 & 0.02 \\ \hline
A-RNN & 0.19 & 0.27 & 0.28 & \textbf{0.38} & 0.19 & 0.02 \\ \hline
AC-RNN & \textbf{0.22} & \textbf{0.33} & \textbf{0.31} & 0.27 & 0.22 & 0.02 \\ \hline
NB & 0.19 & 0.28 & 0.25 & 0.24 & 0.15 & 0.01 \\ \hline
SVM & 0.20 & 0.27 & 0.22 & 0.26 & 0.19 & 0.01 \\ \hline
LSA & 0.20 & 0.30 & 0.24 & 0.30 & 0.17 & 0.02 \\ \hline
LexRank & 0.20 & 0.28 & 0.25 & 0.24 & 0.18 & 0.02 \\ \hline
TextRank & 0.20 & 0.31 & 0.23 & 0.28 & \textbf{0.24} & 0.02 \\ \hline \hline
\multicolumn{7}{|c|}{Target snippet length = 1 sentence} \\ \hline
Fuzzy & 0.11 & 0.15 & 0.18 & 0.25 & 0.09 & 0.003  \\ \hline
E-RNN & 0.12 & 0.15 & \textbf{0.23} & 0.28 & 0.10 & 0.004  \\ \hline
EC-RNN & \textbf{0.13} & \textbf{0.17} & 0.22 & \textbf{0.29} & 0.11 & 0.004  \\ \hline
CNN-RNN & 0.12 & 0.15 & 0.19 & 0.29 & 0.10 & 0.004  \\ \hline
A-RNN & \textbf{0.13} & 0.16 & 0.22 & \textbf{0.29} & 0.11 & 0.004  \\ \hline
AC-RNN & \textbf{0.14} & \textbf{0.19} & \textbf{0.24} & \textbf{0.30} & \textbf{0.11} & \textbf{0.006}  \\ \hline
NB & 0.12 & 0.15 & 0.21 & 0.25 & 0.11 & 0.004  \\ \hline
SVM & 0.12 & 0.15 & 0.20 & 0.27 & 0.11 & 0.004  \\ \hline
LSA & 0.12 & 0.16 & 0.21 & 0.26 & 0.10 & 0.005  \\ \hline
LexRank & 0.12 & 0.15 & 0.19 & 0.25 & 0.10 & 0.004  \\ \hline
TextRank & 0.11 & 0.15 & 0.20 & 0.24 & 0.10 & 0.003  \\ \hline
\end{tabular}
\end{table}
\egroup

\bgroup
\def\arraystretch{0.8}
\begin{table}
\centering
\caption{Results on supervised task using human extracted summaries - 15K
for training and 5K for evaluation. 2 variations
with target summary: When target summaries are extremely long (5
sentence). Extractive Context RNN (EC-RNN)
performs best. Adding context as a whole created improvements in RNN models.}
\label{tab:goldresult5}
\begin{tabular}{|c|c|c|c|c|c|c|} \hline
\multirow{2}{*}{Model} & Token  & \multirow{2}{*}{$Rouge_1$} & \multirow{2}{*}{$Rouge_2$} & Rouge & \multirow{2}{*}{BLEU}  & Topic  \\ 
 & Sim. & & & -LCS &  & Sim. \\ \hline \hline
\multicolumn{7}{|c|}{Target snippet length = 5 sentence} \\ \hline
Fuzzy & 0.28 & 0.37 & 0.28 & 0.21 & 0.30 & 0.03 \\ \hline
V-RNN & 0.33 & 0.41 & \textbf{0.61} & 0.28 & 0.29 & 0.03 \\ \hline
C-RNN & \textbf{0.40} & \textbf{0.52} & 0.40 & 0.31 & \textbf{0.41} & 0.04 \\ \hline
CNN-RNN & 0.32 & 0.43 & 0.29 & 0.26 & 0.36 & 0.04 \\ \hline
A-RNN & 0.31 & 0.36 & 0.29 & \textbf{0.33} & 0.37 & 0.03 \\ \hline
AC-RNN & 0.36 & 0.45 & 0.31 & 0.28 & 0.37 & 0.04 \\ \hline
NB & 0.31 & 0.40 & 0.42 & 0.28 & 0.18 & 0.03 \\ \hline
SVM & 0.32 & 0.40 & 0.38 & 0.29 & 0.25 & 0.03 \\ \hline
LSA & 0.30 & 0.41 & 0.40 & 0.28 & 0.21 & 0.03 \\ \hline
LexRank & 0.30 & 0.40 & 0.30 & 0.28 & 0.28 & 0.03 \\ \hline
TextRank & 0.31 & 0.43 & 0.39 & 0.27 & 0.31 & 0.03 \\ \hline
\end{tabular}
\end{table}
\egroup

\bgroup
\def\arraystretch{1}
\begin{table}
\centering
\caption{Classification for Extractive Supervised Model (Classes: 0 -> Non-summary). 
Training - 15K human judged, Evaluation-5K human judged.
Extractive Context RNN (EC-RNN) shows best performance. Note: No classification for abstractive models.}
\label{tab:goldprecrecall}
\begin{tabular}{|c|c|c|c|c|} \hline
\multirow{2}{*}{Model} & \multirow{2}{*}{Accuracy} & Precision & Recall & F-score \\ 
& & (0,1) & (0,1) & (0,1) \\ \hline
NB & 94.11 & 95, 92.32 & 96, 90 & 95.5, 91.4 \\ \hline
SVM & 93.31 & 97.13, 87.18 & 92.4, 94.91 & 94.74, 90.90 \\ \hline
E-RNN & 95.4 & 98.42, 90.37 & 94.47, 97.12 & 96.40, 93.62 \\ \hline
CNN-RNN & 96.01 & 96.22, 95.56 & 97.69, 92.81 & 96.94, 94.16 \\ \hline
EC-RNN & \textbf{98.1} & \textbf{99.0, 96.07} & \textbf{97.85, 98.27} & \textbf{98.42, 97.16} \\
\hline\end{tabular}
\end{table}
\egroup

\bgroup
\def\arraystretch{0.8}
\begin{table}
\centering
\caption{Supervised model ranking evaluation.  Training - 15K human judged, Evaluation-5K human judged.
Extractive Context RNN (EC-RNN) shows best performance.}
\label{tab:goldranking}
\begin{tabular}{|c|c|c|c|c|} \hline
Model & NDCG@1 & NDCG@3 & MAP@1 & MAP@3 \\ \hline
Fuzzy & 0.535 & 0.576 & 0.087 & 0.144 \\ \hline
E-RNN & 0.617 & 0.636 & 0.094 & 0.159  \\ \hline
EC-RNN & \textbf{0.642} & \textbf{0.682} & 0.102 & \textbf{0.162}  \\ \hline
CNN-RNN & 0.615 & 0.658 & 0.090 & 0.151  \\ \hline
A-RNN & 0.630 & \textbf{0.679} & \textbf{0.109} & 0.155  \\ \hline
AC-RNN & 0.635 & 0.667 & \textbf{0.116} & \textbf{0.160}  \\ \hline
NB & 0.619 & 0.611 & 0.092 & 0.151 \\ \hline
SVM & 0.620 & 0.646 & 0.096 & 0.156  \\ \hline
LSA & \textbf{0.639} & 0.651 & 0.096 & 0.153  \\ \hline
LexRank & 0.607 & 0.644 & 0.095 & 0.153  \\ \hline
TextRank & 0.629 & 0.645 & 0.095 & 0.153  \\ \hline
\end{tabular}
\end{table}
\egroup

\subsection{Results for Supervised Setting}
In this section we present the result for supervised models for summarization. For this puproses we trained our models on 15K human extracted summaries
and evaluated on 5K human extracted summaries. 

Table \ref{tab:goldresult} compiles the performance result for all summarization strategies. It is clear that
RNN's as a whole outperform other techniques (Naive Bayes, SVM, LSA, LexRank, TextRank) on Rouge, BLEU and token similarity.
In both abstractive and extractive RNN's adding document-context improves all metrics. 
For example, Rouge-1 and BLEU for abstractive RNN is 0.27 and 0.19 respectively. When document context is added (AC-RNN) then
Rouge-1 and BLEU increases to 0.33 and 0.22 respectively.
Similarly for Extractive RNN (E-RNN) vs extractive contextual RNN (EC-RNN) rouge-1 changes from 0.27 to 0.39 and BLEU changes from 0.16 to 0.25.

For large target summaries which are 3 sentence long \emph{extractive contextual RNNs} perform the best followed by \emph{abstractive contextual RNNs}. 
For example rouge-1 for EC-RNN is 0.39 and for AC-RNN is 0.33. Similarly BLEU is 0.25 compared to 0.22 for AC-RNN.
For small target summaries (1 sentence long) we observe that abstractive-context-rnn (AC-RNN) outperform extractive-context-rnn (EC-RNN).
Where rouge-1 is 0.15 for EC-RNN compared to 0.19 for AC-RNN.
For target summaries which are 5 sentences long results are shared in Table \ref{tab:goldresult5}. As the target summary length increases efficacy of extractive contextual RNN increases over abstractive contextual RNNs.

In Table \ref{tab:goldprecrecall} we present classification metrics for extractive models.
All models are doing a good job of separating summary-sentences from non-summary sentences. 
For this task EC-RNN outperforms other methodologies as well.

Table \ref{tab:goldranking} describe ranking metrics for summarization models. We assign summary sentences as relevance of 1 and other sentences relevance score of 0.
The task is to generate ranking of sentences in a way that picks the summary sentences before non-summary sentences. 
For this task as well context aware RNNs win out. Extractive context RNN (EC-RNN) have the highest NDCG@1, NDCG@3 and MAP@3.

\subsection{Results for Semi-supervised Setting}
In this section we share the result of training with algorithmically generated approximate summaries. 
These approximate summaries are generated by using document-context. For extractive RNN and other classification 
approaches like SVM, NB human labeled and algorithmically labeled data was used for training. 
Whereas for abstractive RNN we use title and subtitles for learning. 

Table \ref{tab:unsupervisedprecrecall} compares the classification metric for extractive models when trained with
human extracted summaries to algorithmically labeled approximate summaries. It can be seen that training models on 
large scale approximate summaries does not lead to any drop in precision-recall and accuracy.

Table \ref{tab:unsupervisedresult} compares the summarization results for all RNN models (abstractive vs extractive).
Training on 100K approximate summaries does not lead to drop in metric. For example Rouge-1 drops from 0.39 (EC-RNN supervised)
to 0.38 (EC-RNN semi-supervised). BLEU also remains comparable with 0.26 for 100K-dataset and 0.25 for 15-K dataset.

Ranking metrics are compared in Table \ref{tab:unsupervisedranking}.  As the size of data increases for training 
abstractive methods performance increases tremendously. For example AC-RNN have NDCG@3 of 0.667 vs 0.715 for 15K vs 100K documents respectively.
Same result hold for MAP as well, where MAP@3 increase from 0.16 to 0.17 when data is increased from 15K to 100K for AC-RNN.
For extractive RNN, which are trained on approximate summaries we don't see a drop in ranking metrics between 15K human extracted vs 100K approximate summaries.
MAP@3 remains 0.16 in both cases.

Overall it can be seen that abstractive RNNs are improving with more data. 
Extractive RNN are able to generate near similar performance with large-scale approximate summaries as with small scale human extracted summaries. 
And adding document-context to RNN with approximate summaries further boost the performance for both abstractive and extractive RNN.

\bgroup
\def\arraystretch{1}
\begin{table}
\centering
\caption{Classification for Semi-supervised Extractive Supervised model (Classes: 0-> Non-summary). 
Training - 100K algorithmically labeled data, Evaluation-5K human judged.
EC-RNN shows best performance.}
\label{tab:unsupervisedprecrecall}
\begin{tabular}{|c|c|c|c|c|} \hline
\multirow{2}{*}{Model} & \multirow{2}{*}{Accuracy} & Precision & Recall & F-score \\ 
& & (0,1) & (0,1) & (0,1) \\ \hline
\multicolumn{5}{|c|}{Semi-supervised} \\ \hline
E-RNN & 96.77 & 99.37, 92.39 & 95.64, 98.86 & 97.47, 95.52 \\ \hline
CNN-RNN & 97.9 & 97.36, 98.93 & 99.41, 95.22 & 98.38, 97.04 \\ \hline
EC-RNN & \textbf{99.41} & \textbf{99.94, 98.43} & \textbf{99.14, 99.90} & \textbf{99.54, 99.16} \\ \hline
\multicolumn{5}{|c|}{Supervised} \\ \hline
E-RNN & 95.4 & 98.42, 90.37 & 94.47, 97.12 & 96.40, 93.62 \\ \hline
CNN-RNN & 96.01 & 96.22, 95.56 & 97.69, 92.81 & 96.94, 94.16 \\ \hline
EC-RNN & \textbf{98.1} & \textbf{99.0, 96.07} & \textbf{97.85, 98.27} & \textbf{98.42, 97.16} \\ \hline
\end{tabular}
\end{table}
\egroup

\bgroup
\def\arraystretch{0.8}
\begin{table}
\centering
\caption{Result of summarization  for Semi-supervised Extractive Supervised Model. 
Training - 100K algorithmically labeled data, Evaluation-5K human judged.
Extractive Context RNN (EC-RNN) shows best performance. Abstractive Contextual RNNs show significant improvements with large training data.}
\label{tab:unsupervisedresult}
\begin{tabular}{|c@{\hskip1pt}|c@{\hskip1pt}|c@{\hskip1pt}|c@{\hskip1pt}|c@{\hskip1pt}|c@{\hskip1pt}|c@{\hskip1pt}|} \hline
\multirow{2}{*}{Model} & Token  & \multirow{2}{*}{$Rouge_1$} & \multirow{2}{*}{$Rouge_2$} & Rouge & \multirow{2}{*}{BLEU}  & Topic  \\ 
 & Sim. & & & -LCS &  & Sim. \\ \hline \hline
\multicolumn{7}{|c|}{Semi-supervised} \\ \hline
E-RNN & 0.21 & 0.28 & \textbf{0.37} & 0.28 & 0.16 & 0.015  \\ \hline
EC-RNN & \textbf{0.27} & \textbf{0.38} & 0.23 & 0.30 & \textbf{0.26} & \textbf{0.023}  \\ \hline
CNN-RNN & 0.21 & 0.30 & 0.21 & 0.28 & 0.22 & 0.018  \\ \hline
A-RNN & 0.21 & 0.27 & 0.29 & \textbf{0.41} & 0.21 & 0.016  \\ \hline
AC-RNN & 0.23 & 0.33 & 0.32 & 0.29 & 0.22 & 0.021  \\ \hline
\multicolumn{7}{|c|}{Supervised} \\ \hline
E-RNN & 0.20 & 0.27 & \textbf{0.37} & 0.27 & 0.16 & 0.02 \\ \hline
EC-RNN & \textbf{0.26} & \textbf{0.39} & 0.22 & 0.30 & \textbf{0.25} & 0.02 \\ \hline
CNN-RNN & 0.21 & 0.29 & 0.21 & 0.27 & 0.21 & 0.02 \\ \hline
A-RNN & 0.19 & 0.27 & 0.28 & \textbf{0.38} & 0.19 & 0.02 \\ \hline
AC-RNN & 0.22 & 0.33 & 0.31 & 0.27 & 0.22 & 0.02 \\ \hline
\end{tabular}
\end{table}
\egroup

\bgroup
\def\arraystretch{1}
\begin{table}
\centering
\caption{Ranking metrics for Semi-supervised Extractive Supervised model. 
Training - 100K algorithmically labeled data, Evaluation-5K human judged.
Abstractive contextual RNNs show most improvements with large training data. Extractive contextual RNNs also show improvements.}
\label{tab:unsupervisedranking}
\begin{tabular}{|c@{\hskip3pt}|c@{\hskip3pt}|c@{\hskip3pt}|c@{\hskip3pt}|c@{\hskip3pt}|} \hline
Model & NDCG@1 & NDCG@3 & MAP@1 & MAP@3 \\ \hline
\multicolumn{5}{|c|}{Semi-supervised, Supervised} \\ \hline
E-RNN & 0.625, 0.617 & 0.659, 0.636 & 0.097,  0.094 & 0.158, 0.159    \\ \hline
EC-RNN & 0.655, \textbf{0.642} & 0.708, \textbf{0.682} & 0.103, 0.102  & 0.167, \textbf{0.162}   \\ \hline
CNN-RNN & 0.622, 0.615 & 0.676, 0.658 & 0.090, 0.090 & 0.154, 0.151   \\ \hline
A-RNN & 0.649, 0.630 & 0.674, 0.679 & 0.116, 0.109  & 0.160, 0.155   \\ \hline
AC-RNN & \textbf{0.677}, 0.635 & \textbf{0.715}, 0.667  & \textbf{0.125}, \textbf{0.116} & \textbf{0.170}, 0.160  \\ \hline
\end{tabular}
\end{table}
\egroup

\section{Conclusion} 

We proposed a novel Document-context Seq2Seq model for abstractive and extractive summarization. We have shown that RNNs and other Seq2Seq models are powerful and beat state-of-the-art summarization approaches in e-Commerce setting. The idea of adding contextual information at the first time-step during the encoding of the input to output sequence/label mapping aligns with humans since generally humans tend to read title, abstract and gather other contextual information before reading the entire document/articles. This gives humans high-level understanding of the document, which if incorporated in Seq2Seq model will generate much richer document specific summaries. Training is performed in a human tagged supervised setting as well as with Large-Scale-Semi-Supervised extracted summaries. It is found that Seq2Seq based RNN  summarization techniques out-performs other state-of-the-art summarization techniques. Within RNNs, Contextual-RNNs outperform Non-Contextual-RNNs on most of the similarity and ranking measures. Extractive-Contextual RNNs are found to be best performing followed by Abstractive-Contextual RNNs for large summaries, however for shorter summaries Abstractive-Contextual-RNNs outperform all other techniques followed by Extractive-RNNs, with Attention and more sophisticated setting it can be possible to further improve Abstractive techniques. We have also depicted that Abstractive-RNNS (Contextual/Non-Contextual) can be used for Extraction tasks and still beat the Extractive systems. It is found that Large-Scale-Semi-Supervised data for training improves the performance of the models on the golden-evaluation-dataset. Hence, training the extractive models with approximate summaries leads to better results compared to relatively smaller human tagged supervised data. We think that advantage of large scale training outperforms the noise in approximating summaries. We recommend other researchers to incorporate context in other Seq2Seq tasks (e.g. machine translation task). 


%
\bibliographystyle{abbrv}
\bibliography{sigproc}  
%
%
\end{document}